\PassOptionsToPackage{table,x11names}{xcolor}
\documentclass[runningheads]{llncs}
\usepackage[T1]{fontenc}
\usepackage{orcidlink}
\usepackage[utf8]{inputenc}

\usepackage[top=80pt,bottom=80pt,left=88pt,right=86pt]{geometry}
\usepackage{graphicx}
\usepackage{array}
\usepackage{algorithm}
\usepackage[noend]{algorithmic}
\usepackage{microtype}
\usepackage{subfig}
\usepackage{tabularx}

\usepackage{booktabs}
\usepackage{amsmath}

\usepackage{enumerate}
\usepackage{tikz}
\usetikzlibrary{shapes.geometric}
\usepackage{wrapfig}
\usepackage{multirow}
\usepackage{nicefrac}
\usepackage[colorinlistoftodos,prependcaption,textsize=tiny]{todonotes}
\usepackage[numbers,square,sort]{natbib}

\usepackage{url}

\def\eps{{\varepsilon}}

\begin{document}

\title{I See Dead People:\\Gray-Box Adversarial Attack on Image-To-Text Models\thanks{Supported by the Israeli Innovation Authority through the Trust.AI consortium, and the Lynne and William Frankel Center for Computer Science.}}

\titlerunning{I See Dead People}

\author{Raz Lapid\,\orcidlink{0000-0002-4818-9338}\inst{1,2} \and
Moshe Sipper\orcidlink{0000-0003-1811-472X}\inst{1}}

\authorrunning{Lapid \& Sipper}

\institute{Department of Computer Science, Ben-Gurion University of the Negev, Beer-Sheva, 8410501, Israel
\and
DeepKeep, Tel-Aviv, Israel \\
\email{sipper@bgu.ac.il}\\
\url{http://www.moshesipper.com/}}

\maketitle

\begin{abstract}
Modern image-to-text systems typically adopt the encoder-decoder framework, which comprises two main components: an image encoder, responsible for extracting image features, and a transformer-based decoder, used for generating captions. Taking inspiration from the analysis of neural networks' robustness against adversarial perturbations, we propose a novel gray-box algorithm for creating adversarial examples in image-to-text models. Unlike image classification tasks that have a finite set of class labels, finding visually similar adversarial examples in an image-to-text task poses greater challenges because the captioning system allows for a virtually infinite space of possible captions. In this paper, we present a gray-box adversarial attack on image-to-text, both untargeted and targeted. We formulate the process of discovering adversarial perturbations as an optimization problem that uses only the image-encoder component, meaning the proposed attack is language-model agnostic. Through experiments conducted on the ViT-GPT2 model, which is the most-used image-to-text model in Hugging Face, and the Flickr30k dataset, we demonstrate that our proposed attack successfully generates visually similar adversarial examples, both with untargeted and targeted captions. Notably, our attack operates in a gray-box manner, requiring no knowledge about the decoder module. We also show that our attacks fool the model hosted in the popular open-source platform Hugging Face.
\end{abstract}

\begin{figure}
    \centering
    \begin{tabular}{cccc}
        \subfloat[(Orig) A man and a woman posing for a picture] {\includegraphics[scale=0.4]{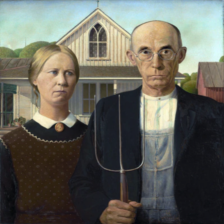}} \hspace{0.3cm} &
        \subfloat[(Adv) A table with a bunch of stuffed animals on top of it]{\includegraphics[scale=0.4]{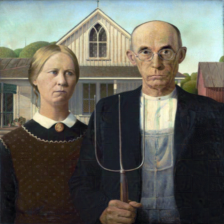}} \hspace{1cm}
        \subfloat[(Orig) A woman sitting on a chair in a room]{\includegraphics[scale=0.4]{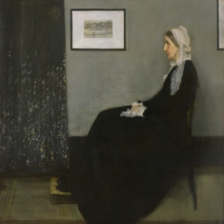}} \hspace{0.3cm} &
        \subfloat[(Adv) Two men standing next to each other near a truck]{\includegraphics[scale=0.4]{figures/paintings/pert_1.png}} 
    \end{tabular}
    \caption{Original and adversarial paintings of the well-known painting ``American Gothic'' by Grant Wood and ``Whistler's Mother'' by James Abbott McNeill Whistler. Left: Original image and its caption (Orig); right: adversarial image and its caption (Adv).}
\end{figure}


\section{Introduction}
\label{sec:intro}
Rapid progress in deep learning and artificial intelligence over the past few years has led to remarkable advances in various domains, including computer vision and natural language processing \cite{vaswani2017attention,radford2021learning,ramesh2021zero,saharia2022photorealistic,radford2018improving}. One significant area of research lies at the intersection of these two fields---image-to-text models. These models aim to bridge the gap between visual and textual information modalities by generating descriptive and accurate textual descriptions of images. They have proven to be valuable in applications such as image captioning \cite{li2022blip,li2023blip}, visual question answering \cite{antol2015vqa,kim2021vilt}, and image retrieval \cite{lu2019vilbert,cao2018vggface2}.

Despite their impressive, performance image-to-text models are not free of problematic vulnerabilities. One significant vulnerability, which has garnered substantial attention, is susceptibility to adversarial attacks \cite{goodfellow2014explaining,szegedy2013intriguing}. Adversarial attacks are carefully crafted input perturbations designed to deceive machine learning models \cite{lapid2023patch,tamam2022foiling,zhuang2023pilot,carlini2017towards,brown2017adversarial,lapid2022evolutionary}. These perturbations are imperceptible to human observers---but can provoke incorrect or unintended outputs from the targeted model.

Adversarial attacks on image-to-text models can have severe implications, because they can compromise the reliability and trustworthiness of these models in real-world scenarios. An attacker can manipulate images in subtle ways, thus causing the model to generate misleading or even malicious textual descriptions. These attacks have significant consequences in various applications, including autonomous driving systems, assistive technologies for the visually impaired, and content-moderation systems.

The AI community has witnessed a significant trend towards the utilization of pretrained models in recent years \cite{kolesnikov2020big,li2023blip}. Pretrained models are deep learning models that have been trained on huge datasets and complex tasks, such as image classification, natural language processing, or even multimodal tasks like image-and-text generation. These models learn rich representations of data and capture valuable knowledge that can be transferred to new tasks.

However, pretrained models introduce the risk of being attacked in a query-free manner \cite{zhuang2023pilot}. Traditional adversarial attacks often involve crafting perturbations or input samples specifically designed to deceive a model when it receives a particular query or input. However, query-free attacks target the model's vulnerability without requiring direct access to the overall end-to-end pipeline.

\begin{figure}
    \centering
    \includegraphics[scale=0.25]{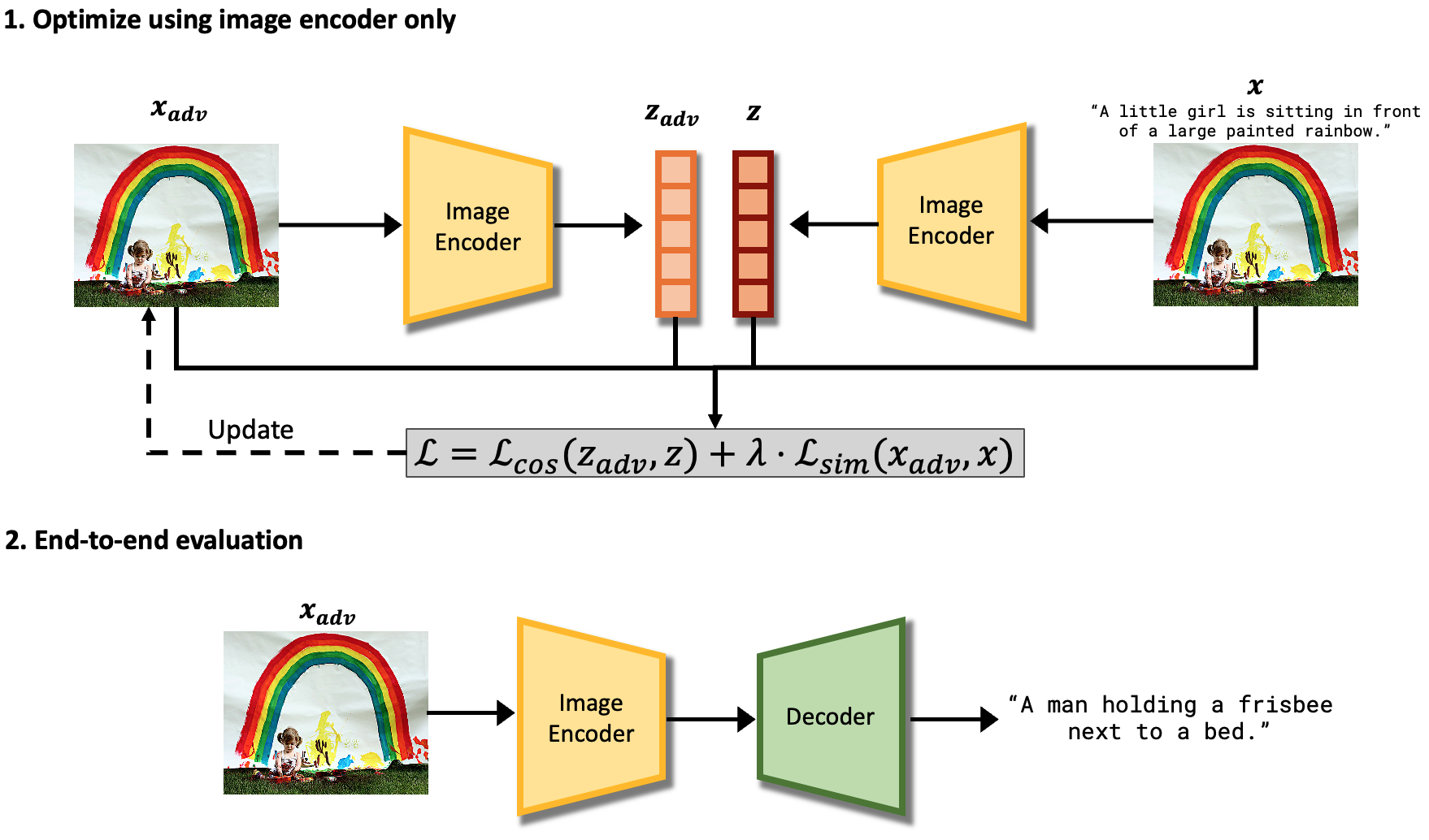}
    \caption{Overview of the attack. First, we feed the original image $x$ to the image encoder, to get the embedding $z$. Then we generate a perturbation $x_{adv}$ using only the image encoder. Finally, once we have finished the optimization process, we evaluate on the end-to-end pipeline.}
    \label{fig:overview}
\end{figure}

Focusing only on the image encoder, this paper proposes a gray-box adversarial attack on image-to-text models---both untargeted and targeted.
Specifically, our contributions are:
\begin{enumerate}
    \item A gray-box, bounded attack that does not query the end-to-end pipeline (Figure~\ref{fig:overview}).
    
    \item Empirical demonstration of the effectiveness of our novel attack---both in untargeted and targeted scenarios.
    
    \item Demonstration that the attack is effective also while using the Hugging Face hosted inference API (Figure~\ref{images:huggingface}).
\end{enumerate}

\section{Previous Work}
\label{sec:prev}
\textbf{Adversarial attacks.} Adversarial attacks manipulate Deep Neural Networks (DNNs) by introducing carefully-designed small perturbations into input data \cite{madry2017towards,croce2020reliable,xu2018structured,tamam2022foiling,ilyas2018black,xiao2018spatially,lapid2022evolutionary,wang2021similarity,aafaq2022language,goodfellow2014explaining,lapid2023patch,andriushchenko2020square}. These attacks can be classified into three types, based on how the attacker interacts with the victim model: (1) white-box attacks, where the attacker has full access to the victim model and exploits local gradient information to generate attacks \cite{madry2017towards,croce2020reliable,xiao2018spatially}; (2) black-box attacks, where the attacker only has access to the input and output of the victim model and employs input-output queries to generate attacks, such as score-based attacks and decision-based attacks \cite{lapid2022evolutionary,andriushchenko2020square,lapid2023patch,ilyas2018black}; (3) and gray-box attacks, which lie between white-box and black-box attacks, where the attacker possesses partial knowledge or limited access to the victim model \cite{aafaq2022language,wang2021similarity}. 

In this study, we consider a scenario where the adversary has access to the vision transformer (ViT) \cite{vaswani2017attention} image encoder but lacks knowledge of the end-to-end image-to-text model used for caption generation. Our objective is to develop an adversarial attack that can deceive the image-to-text model without actually running the computationally expensive captioning process or requiring extensive model queries. Hence, this attack is gray-box.

\textbf{Adversarial attacks in vision-language models.} Prior studies have delved into the realm of adversarial attacks on image-to-text models, aiming to uncover vulnerabilities and explore potential defenses. Researchers have investigated various techniques and attack vectors to evaluate the robustness of these models. 

Chen et al. \cite{chen2017show} proposed a method for generating both untargeted and targeted perturbations on image inputs to manipulate the output text captions, showcasing the potential for adversarial attacks in this domain. However, they focused on a white-box threat model and convolutional neural networks (CNNs) \cite{krizhevsky2017imagenet}, which were the dominant architecture in computer vision at the time. 

\cite{aafaq2022language} proposed an attack on image-to-text models using GANs. This approach is much more computationally expensive since they trained the GAN to generate perturbations as part of the optimization. GANs are trained in a min-max game between a generator and a discriminator network. This adversarial training process can lead to unstable dynamics, commonly known as mode collapse or oscillation. 

Similarly, Zhuang et al, \cite{zhuang2023pilot} introduced an approach to crafting adversarial examples on a text prompt for Stable Diffusion \cite{rombach2022high}, even in the absence of end-to-end model queries. 

These studies shed light on the susceptibility of vision-language models to adversarial attacks, emphasizing the need for improved robustness and security measures in this field. 

\section{Our Proposal}
\label{sec:proposal}
\textbf{Problem statement.} In this section we initially provide a comprehensive overview of the image-to-text task, followed by the introduction of our primary objective: the generation of subtle perturbations on visual inputs to strategically influence the synthesized captions produced by the model.

We specifically chose the ViT-GPT2 image-captioning model \cite{nlp_connect_2022}, which is the most commonly used image-to-text model on Hugging Face. ViT-GPT2 is a powerful deep learning model that combines the capabilities of a vision transformer \cite{dosovitskiy2020image} and GPT-2 \cite{radford2019language} to perform image-to-text generation tasks. This model is designed to understand and generate textual descriptions or captions based on input images. 

The ViT component of the model is responsible for processing the visual information contained in the images. It applies a self-attention mechanism \cite{zhao2020exploring} to capture meaningful relationships between different image patches, allowing it to understand the content and context of the visual data. The GPT-2 component is a language model that excels at generating coherent and contextually relevant text. It uses a transformer architecture to capture the dependencies between words and generate human-like text, based on the learned patterns and relationships in the training data. 

By combining these two components, the image-to-text ViT-GPT2 model can effectively bridge the gap between images and text. It takes an image as input, passes it through the vision transformer (image encoder) to extract visual features, and then feeds those features into the GPT-2 model (decoder) to generate textual descriptions or captions that correspond to the image.

Herein, we focus on the image encoder, ViT; thus, the overall pipeline isn't aware of the attack. As depicted in Figure~\ref{fig:overview}, we generate small perturbations on the image input of the image encoder to cause the embedding to change. We make use of cosine similarity (Equation~\ref{eq:cosine}), which measures similarity between two vectors (value $\in [-1, 1]$, -1 least similar, +1 most similar).

\begin{equation}
    CS(u, v) = \frac{u\cdot v}{\Vert u \Vert \cdot \Vert v \Vert}.
    \label{eq:cosine}
\end{equation}

\textbf{Threat model.} We assume that the adversary possesses the trained image encoder of the end-to-end model, granting them the ability to manipulate the input image using the trained image-to-text model. This manipulation involves generating noise that is bounded by $\Vert \cdot \Vert_{\infty}$. Let $f_{\theta}(x)$ represent the ViT image encoder with parameters $\theta$ when evaluated on the image input $x$. We refer to the perturbed input image used as the input to ViT-GPT2 as $x_{adv}$. The attacker's objective is defined as minimizing the cosine similarity between the image embeddings of $x$ and $x_{adv}$ while keeping the perturbed image similar to the original image. Consequently, this gives rise to the attack-generation problem described as follows:
\begin{equation}
\label{eq:untargeted}
    \mathcal{L}^u_{CS} = CS(f_{\theta}(x), f_{\theta}(x_{adv})),
\end{equation}
where $CS$ refers to the cosine similarity metric, and $u$ refers to the attack being untargeted. Minimizing this objective favors a dissimilar embedding vector, $f_{\theta}(x_{adv})$. 

In the targeted scenario, we use the following objective:
\begin{equation}
\label{eq:targeted}
    \mathcal{L}^t_{CS} = 1 - CS(f_{\theta}(x_{target}), f_{\theta}(x_{adv})),
\end{equation}
where $t$ refers to the attack being targeted, and $x_{target}$ is the target image. This objective will result in $ f_{\theta}(x_{adv}) \approx f_{\theta}(x_{target})$, i.e., the embedding vectors of attack vector and target vector are similar. For the image-similarity constraint, we use $\Vert \cdot \Vert_2$ minimization as follows:
\begin{equation}
\label{eq:sim}
    \mathcal{L}_{sim}=\Vert x - x_{adv} \Vert_2,
\end{equation}

Overall, we get the next objective:
\begin{equation}
\label{eq:obj}
\begin{array}{l}
    \mathcal{L}  = \mathcal{L}_{CS} + \lambda \cdot \mathcal{L}_{sim}, \\
    s.t. \, \Vert x - x_{adv} \Vert_{\infty} < \eps,
\end{array}
\end{equation}

where $\lambda$ is a regularization term and $\eps$ is a bounding term. 
The parameter $\lambda$ is used to balance the contributions of the adversarial loss and the image-similarity loss.
A higher value of $\lambda$ places more emphasis on maintaining image similarity, while a lower value places more emphasis on deceiving the Image-To-Text model. In our experiments we set $\lambda$ to 0.1, based on empirical observations and prior literature.

Essentially, find a minimal perturbation such that the perturbed image is kept similar to the original one, and the embedding vector is dissimilar, thus resulting in a different explanation.

Although the attack generation process is simple, we will demonstrate its effective use in targeting image-to-text. Significantly, the generation of the modified input $x_{adv}$ depends only on the image encoder and not on the end-to-end pipeline, leading to improved computational efficiency.

The $\Vert \cdot \Vert_{\infty}$ norm constraint in Equation~\ref{eq:obj} controls the maximum pixel-wise perturbation allowed in the adversarial image. By setting a threshold on the maximum allowable perturbation, we ensure that the changes made to the image are small enough to be imperceptible to the human eye. This constraint helps us limit the visibility of the perturbation and maintain a high degree of visual similarity between the original and adversarial images.

On the other hand, the $\Vert \cdot \Vert_{2}$ norm constraint in Equation~\ref{eq:sim}  ensures that the overall magnitude of the perturbation is minimized. By minimizing the $\Vert \cdot \Vert_{2}$ distance between $x_{adv}$ and $x$, we encourage small changes across all pixels rather than a few large changes concentrated in specific regions. This constraint helps to preserve the global structure and appearance of the original image, enhancing the visual similarity between $x_{adv}$ and $x$.

\textbf{Attack method.} Considering the differentiability of objective \ref{eq:obj} there is a range of optimization methods available for generating attacks. Drawing inspiration from prior research on adversarial attacks in the vision domain, we propose the following attack method.

We make use of the PGD (projected gradient descent) attack \cite{madry2017towards} to craft the perturbation, with some changes: We use the \texttt{Adam} \cite{kingma2014adam} optimizer with default parameters set by PyTorch \cite{NEURIPS2019_9015}, ($\beta_1 = 0.9, \beta_2 = 0.999$), and a learning rate $0.01$. We use the \texttt{ReduceLROnPlateau} learning rate scheduler that reduces the learning rate by a factor of $0.1$ if it detects a plateau in the optimization process for $30$ successive epochs. We used the above hyperparameters both for untargeted and targeted scenarios.

\section{Experiment}
\label{sec:experiment}
\subsection{Experiment setups}
\textbf{Model setup.} In our experiments, the ViT-GPT2-image-captioning model serves as the victim model for generating captions. The attack we propose operates without explicit queries and leverages the image encoder model, ViT-base model, which has been trained on the ImageNet-21k dataset \cite{ridnik2021imagenet} at a resolution of $224 \times 224$. The decoder, on the other hand, is based on the GPT2 model. 

\textbf{Dataset.} We used the Flickr30k dataset \cite{plummer2015flickr30k}, denoted by $\mathcal{D}$, which is a widely used dataset in computer vision and natural language processing. It contains 31,783 images sourced from Flickr, each paired with five descriptive captions.  At first, we witnessed model hallucinations \cite{ji2023survey} on samples taken from $\mathcal{D}$, so we filtered the data by using the CLIP model (ViT-B/32), denoted by $\mathcal{C}$, \cite{radford2021learning}, which was designed to understand and connect images and their corresponding textual descriptions. 

For a given data sample $(x, y)\in \mathcal{D}$, we fed the target model with $x$ and received a caption prediction $y_{pred}$. Then, we fed both $y$ and $y_{pred}$ to $\mathcal{C}$ and checked the cosine similarity of the given embeddings. We set the threshold to $\tau=0.7$, thus filtering all pairs that had less then $\tau$ cosine similarity with the ground truth.

\textbf{Evaluation metrics.} We randomly picked 1000 images from the dataset, \\ $\{(x, y) \in \mathcal{D} | CS(\mathcal{C}(f_{\theta}(x)), \mathcal{C}(y)) \geq \tau \}$.  We assessed the efficacy of an attack by utilizing the CLIP score to gauge the similarity between the image input and the generated text. A reduced CLIP score signifies a diminished semantic correlation between the generated image and the input text, indicating the increased effectiveness of the attack method. We tested our approach using 4 different $\epsilon$ values, $\epsilon \in \{0.05, 0.1, 0.2, 0.3\}$. We evaluated attack performance using the CLIP score \cite{radford2021learning}: a similarity measure computed by the CLIP model to measure the similarity or relatedness between an image and a textual description. Given an image and a segment of text, CLIP computes a similarity score based on how well the image and text align semantically. The score indicates how closely the visual content of the image matches  the textual information provided. 

\subsection{Experiment results}
\textbf{Untargeted results.} In the untargeted scenario we used Equation~\ref{eq:untargeted} for the optimization process. We used the CLIP score to assess the performance of the proposed attack, shown in Table~\ref{tab:untargeted}. Our experimental results revealed compelling findings regarding the effectiveness of the untargeted attack approach. By applying our algorithm to a diverse range of images, we consistently achieved a high success rate in generating adversarial examples. These adversarial examples, crafted through the untargeted attack, were able to mislead the models into making incorrect predictions without a specific target class. 

\begin{figure}
    \begin{tabular}{cccccc}
        \subfloat[\tiny{(Orig) A man wearing a hat and sunglasses}]{\includegraphics[width = 0.91in]{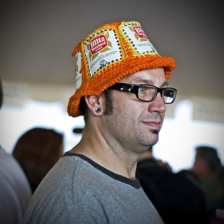}} \;&
        \subfloat[\tiny{(Adv) A woman kneeling down next to a dog}]{\includegraphics[width = 0.91in]{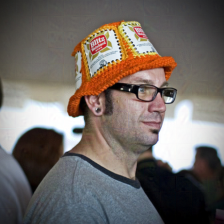}} \;& \quad
        \subfloat[\tiny{(Orig) A man with a dog standing next to a parked car}]{\includegraphics[width = 0.91in]{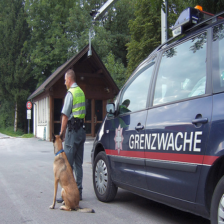}} \;& 
        \subfloat[\tiny{(Adv) A cake sitting on top of a wooden table}] 
        {\includegraphics[width = 0.91in]{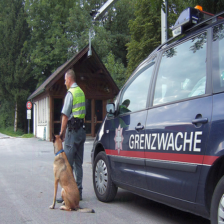}} \;& 
        \quad
        \subfloat[\tiny{(Orig) A man riding a bike down a street}]{\includegraphics[width = 0.91in]{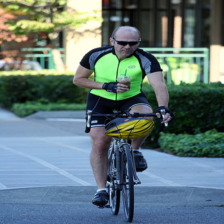}} \;&
        \subfloat[\tiny{(Adv) A man standing next to a woman in front of a fire hydrant}]{\includegraphics[width = 0.91in]{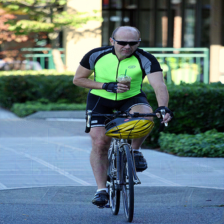}}  \\
        
        \subfloat[\tiny{(Orig) A man sitting on a wooden bench with a guitar}]{\includegraphics[width = 0.91in]{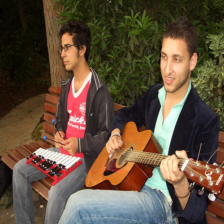}} \;&
        \subfloat[\tiny{(Adv) A woman standing in front of a bunch of bananas}]{\includegraphics[width = 0.91in]{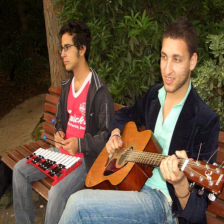}} \;& \quad
        \subfloat[\tiny{(Orig) Two men walking down the street with luggage}]{\includegraphics[width = 0.91in]{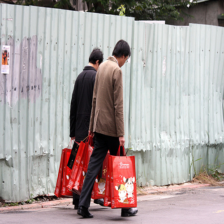}} \;&
        \subfloat[\tiny{(Adv) A woman is playing a video game on a table}]{\includegraphics[width = 0.91in]{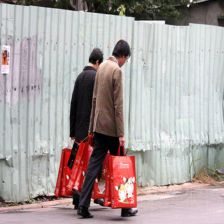}} \;&  \quad
        \subfloat[\tiny{(Orig) A man that is sitting in front of a drum set}]{\includegraphics[width = 0.91in]{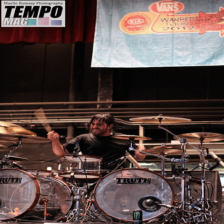}} \;&
        \subfloat[\tiny{(Adv) A man and a woman standing in front of a store}]{\includegraphics[width = 0.91in]{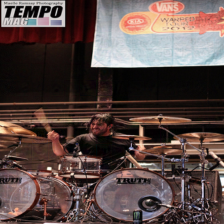}} \\
        
        \subfloat[\tiny{(Orig) A man riding a dirt bike on top of a log}]{\includegraphics[width = 0.91in]{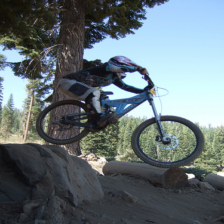}} \;&
        \subfloat[\tiny{(Adv) Two people standing next to each other in front of a fence}]{\includegraphics[width = 0.91in]{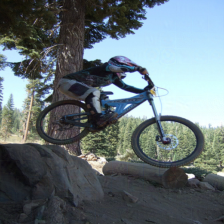}} & \quad
        \subfloat[\tiny{(Orig) Two dogs playing with each other in the snow}]{\includegraphics[width = 0.91in]{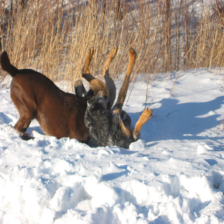}} \;&
        \subfloat[\tiny{(Adv) A woman sitting on top of a bed next to a bird}]{\includegraphics[width = 0.91in]{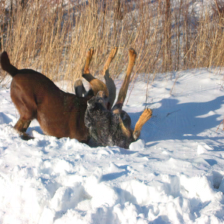}}  \;& \quad
        \subfloat[\tiny{(Orig) A man standing in front of a food cart filled with food}]{\includegraphics[width = 0.91in]{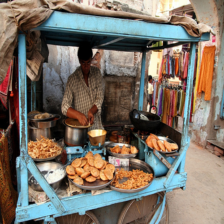}} \;&
        \subfloat[\tiny{(Adv) Three people sitting on a chair with a vase of flowers}]{\includegraphics[width = 0.91in]{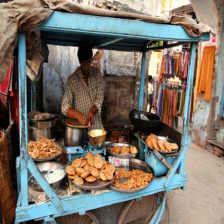}} \\

    \end{tabular}
    \caption{Examples of untargeted gray-box attacks ($\epsilon=0.05$). Pairs of original images and their captions (Orig) beside adversarial images and their captions (Adv).}
    \label{images:untargeted}
\end{figure}

In this experiment we see that increasing $\epsilon$ does not result in lower CLIP scores. We hypothesize that the lack of correlation between increasing perturbation size and lower CLIP scores could be attributed to the fact that excessively large perturbations may introduce noticeable artifacts or distortions, making the adversarial images appear less natural and reducing their similarity to the original images. Consequently, the Image-To-Text models may become more cautious in assigning incorrect descriptions, resulting in higher CLIP scores for these perturbed images. Samples from the experiments are shown in Figure~\ref{images:untargeted}. 

\begin{table}
\centering
\captionof{table}{Results of our algorithm, both for untargeted and targeted attacks: 
Mean performance on 1000 randomly selected images from the Flickr30k dataset.
Values shown are mean $\pm$ std CLIP scores, which are between 0--1. 
The higher the CLIP score between two captions, the more similar the captions are.
No Attack: CLIP scores before the attack of the image with the ground truth caption. 
Our Attack: CLIP scores after our attack. 
In all cases the attack reduces the score, i.e., captions are less similar (with the images themselves retaining high similarity). The lowest (best) score is in bold and the highest difference achieved is highlighted in gray.}
\label{tab:res}
\begin{tabular}{cc}   
    Untargeted & Targeted \\  
    \begin{tabular}{ccc}
        \toprule
        \textbf{$\epsilon$} & \textbf{No Attack} & \textbf{Our Attack}  \\
        \midrule
        $0.05$ & $0.273$ $\pm$ \small{$0.031$} &  $0.166$ $\pm$ \small{$0.034$}\\
        $0.1$ & $0.314$ $\pm$ \small{$0.029$} &  $\boldsymbol{0.155}$ $\pm$ \small{$0.035$}\\
        $0.2$ & $0.279$ $\pm$ \small{$0.029$} &  $0.182$ $\pm$ \small{$0.033$}\\
        \cellcolor{lightgray}
        $0.3$ & \cellcolor{lightgray} $0.364$ $\pm$ \small{$0.030$} &  \cellcolor{lightgray} $0.189$ $\pm$ \small{$0.033$} \\
        \bottomrule
        \label{tab:untargeted}
    \end{tabular} & \quad \quad \quad  
    \begin{tabular}{ccc}
        \toprule
        \textbf{$\epsilon$} & \textbf{No Attack} & \textbf{Our Attack}  \\
        \midrule
        $0.05$ & $0.261$ $\pm$ \small{$0.024$} &  $0.161$ $\pm$ \small{$0.045$} \\
        \cellcolor{lightgray}
        $0.1$ & \cellcolor{lightgray} $0.329$ $\pm$ $0.024$ &  \cellcolor{lightgray} $0.167$ $\pm$ $0.046$ \\
        $0.2$ & $0.301$ $\pm$ \small{$0.024$} &  $0.228$ $\pm$ \small{$0.045$} \\
        $0.3$ & $0.286$ $\pm$ \small{$0.025$} &  $\boldsymbol{0.134}$ $\pm$ $0.045$ \\
        \bottomrule
        \label{tab:targeted}
  \end{tabular} \\
\end{tabular}
\end{table}

\textbf{Targeted results.} In the targeted scenario we used Equation~\ref{eq:targeted} for the optimization process. We observed that the perturbations that were introduced successfully deceived the targeted model. We ensured that every pair of images---Orig and Target---held the $\tau$ constraint. The results are shown in Table~\ref{tab:targeted}. We achieved the lowest CLIP score in the targeted scenario, $0.134$, which shows the efficacy of the proposed attack, without modifying any hyperparameter. By carefully crafting adversarial examples, the model was consistently fooled, often outputting the exact target captions.

\begin{figure}
    \begin{tabular}{cccccc}
        \subfloat[\tiny{(Orig) A store filled with lots of different types of clothing}]{\includegraphics[width = 0.91in]{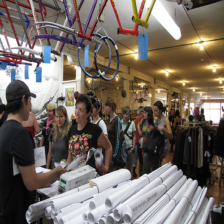}} \;&
        \subfloat[\tiny{(Target) A little boy sitting in a pool of water}]{\includegraphics[width = 0.91in]{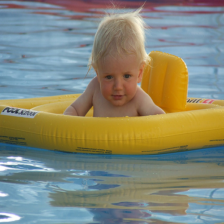}} \;& 
        \subfloat[\tiny{(Adv) A little boy in a swimming pool with a surfboar}]{\includegraphics[width = 0.91in]{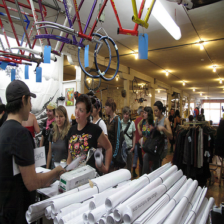}} \;& 
        \quad
        \subfloat[\tiny{(Orig) A man riding a motorcycle on a race track}]{\includegraphics[width = 0.91in]{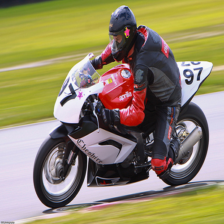}} \;&
        \subfloat[\tiny{(Target) A person riding skis down a snow covered slope}]{\includegraphics[width = 0.91in]{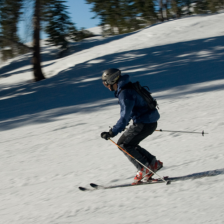}} \;& 
        \subfloat[\tiny{(Adv) A person riding skis down a snow covered slope}]{\includegraphics[width = 0.91in]{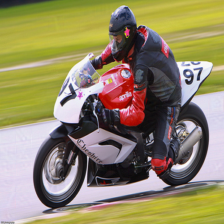}}  \\

        \subfloat[\tiny{(Orig) A person on a skateboard jumping over a wall}]{\includegraphics[width = 0.91in]{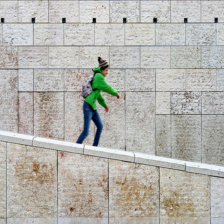}} \;&
        \subfloat[\tiny{(Target) A man paddling a boat down a river}]{\includegraphics[width = 0.91in]{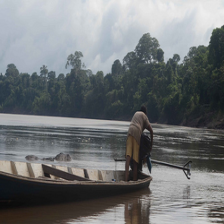}} \;& 
        \subfloat[\tiny{(Adv) A man paddling a canoe down a river}]{\includegraphics[width = 0.91in]{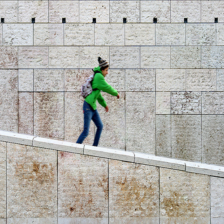}} \;& 
        \quad
        \subfloat[\tiny{(Orig) A baseball player holding a bat on a field}]{\includegraphics[width = 0.91in]{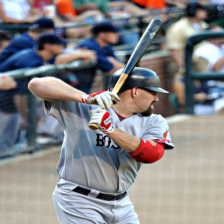}} \;&
        \subfloat[\tiny{(Target) A man riding on the back of a white horse}]{\includegraphics[width = 0.91in]{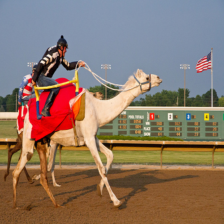}} \;& 
        \subfloat[\tiny{(Adv) A man on a field with a horse}]{\includegraphics[width = 0.91in]{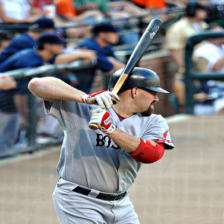}}  \\

        \subfloat[\tiny{(Orig) A woman standing next to a fountain in a city}]{\includegraphics[width = 0.91in]{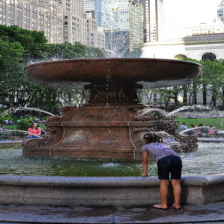}} \;&
        \subfloat[\tiny{(Target) A man riding a wave on top of a surfboard}]{\includegraphics[width = 0.91in]{figures/targeted/0.05/target_1059.png}} \;& 
        \subfloat[\tiny{(Adv) A man riding a wave on top of a surfboard}]{\includegraphics[width = 0.91in]{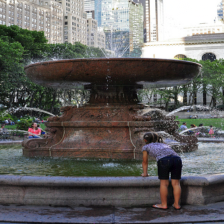}} \;& 
        \quad
        \subfloat[\tiny{(Orig) A man in a suit talking on a microphone}]{\includegraphics[width = 0.91in]{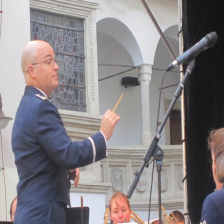}} \;&
        \subfloat[\tiny{(Target) A large group of people standing in front of a building}]{\includegraphics[width = 0.91in]{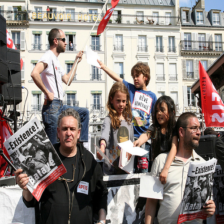}} \;& 
        \subfloat[\tiny{(Adv) A crowd of people standing near a building}]{\includegraphics[width = 0.91in]{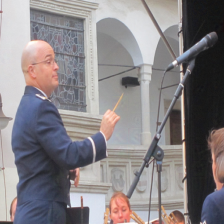}}  \\

    \end{tabular}
    \caption{Examples of targeted gray-box attacks ($\epsilon=0.05$). Triplets of original images and their captions (Orig), target images and their captions (Target), and adversarial images and their captions (Adv).}
    \label{images:targeted}
\end{figure}

These results highlight the vulnerability of machine learning models to targeted adversarial attacks and the need for robust defenses to ensure the reliability and security of AI systems in real-world applications. Samples from the experiments are shown in Figure~\ref{images:targeted}.

\begin{figure}
    \begin{tabular}{cccc}
        \subfloat[(Orig)]{\includegraphics[width = 1.47in]{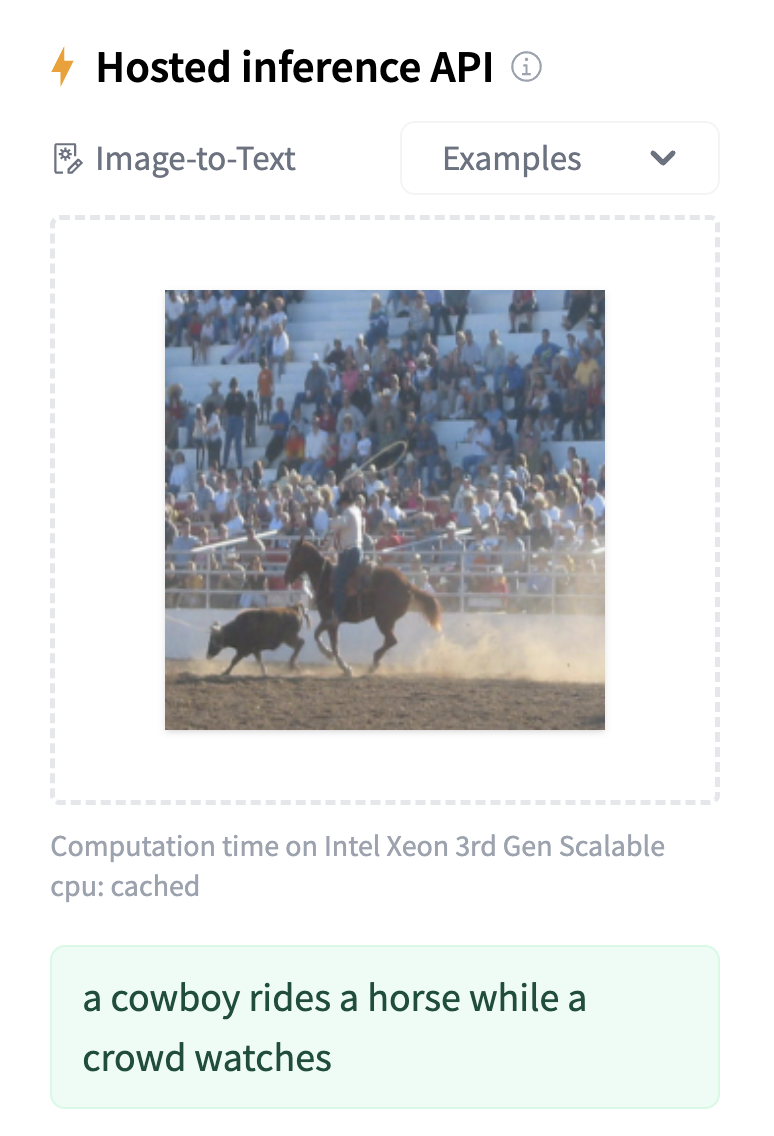}} &
        \subfloat[(Adv)]{\includegraphics[width = 1.47in]{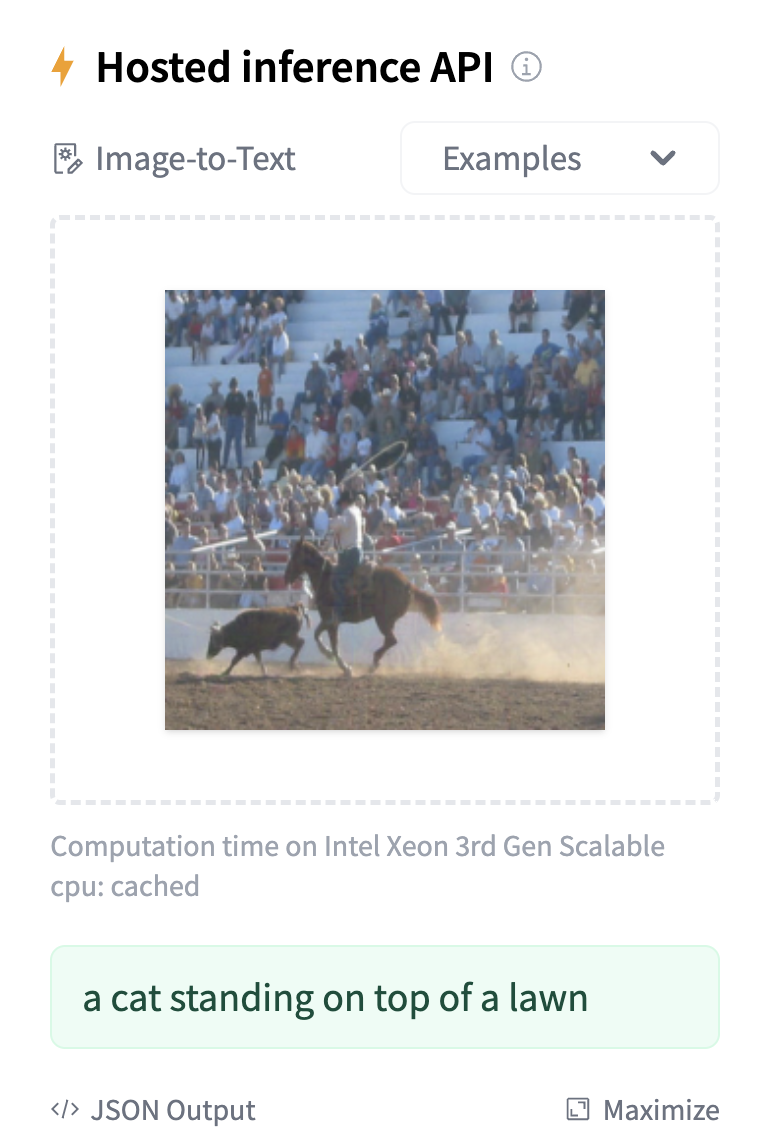}} & \quad
        \subfloat[(Orig)]{\includegraphics[width = 1.47in]{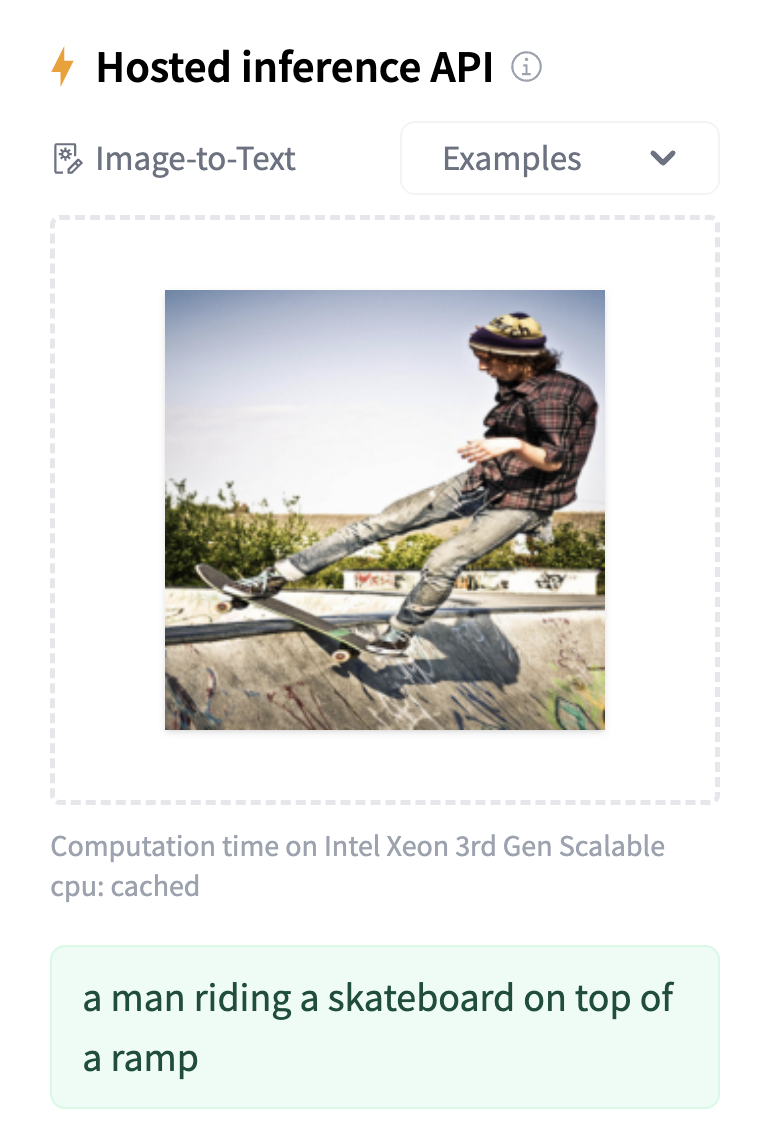}} &
        \subfloat[(Adv)]{\includegraphics[width = 1.47in]{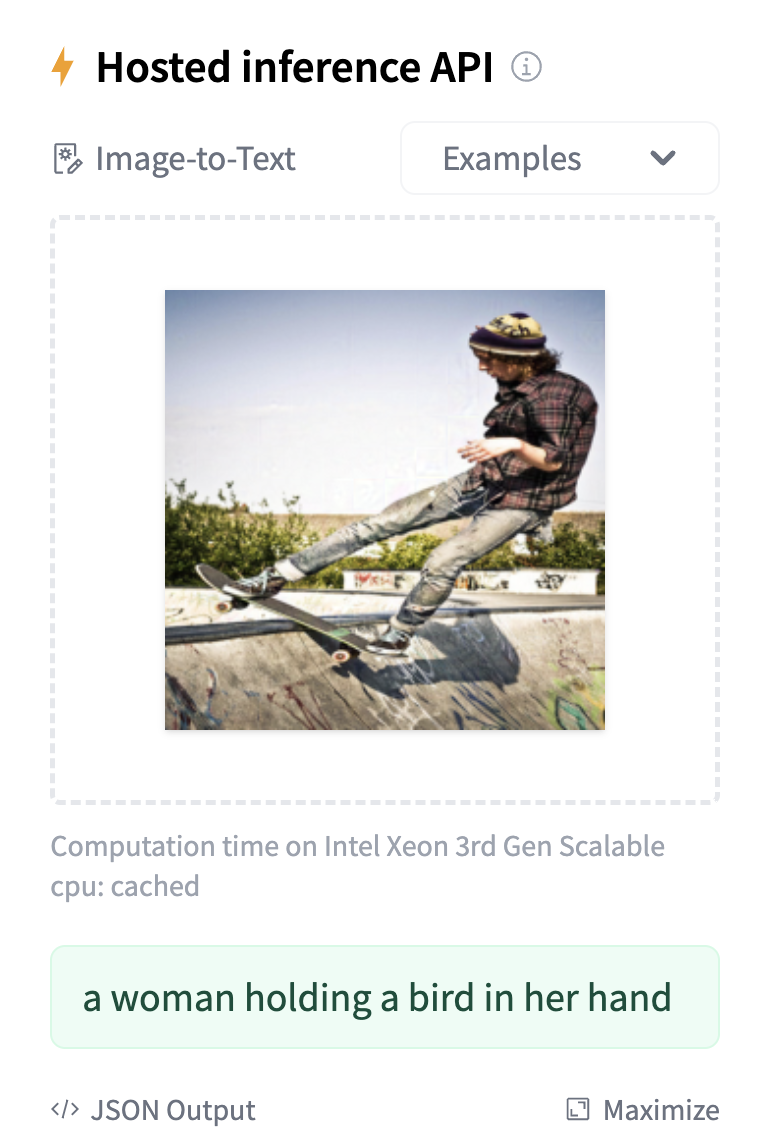}} \\
        
        \subfloat[(Orig)]{\includegraphics[width = 1.47in]{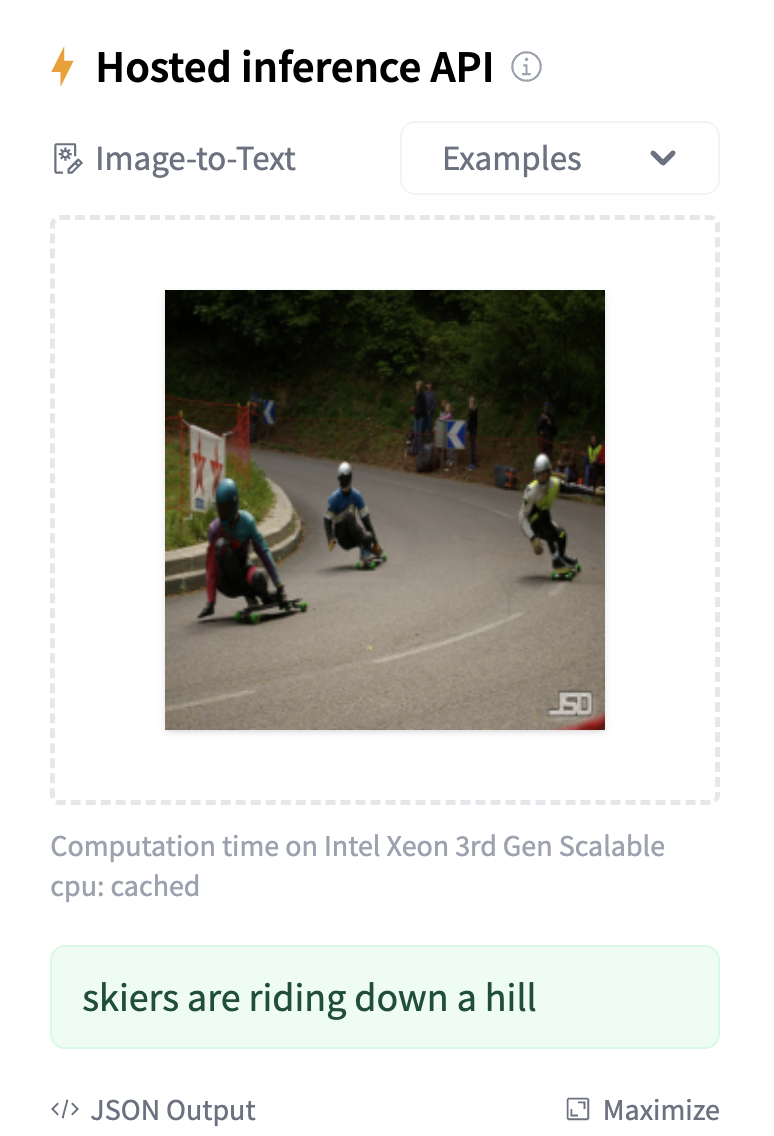}} &
        \subfloat[(Adv)]{\includegraphics[width = 1.47in]{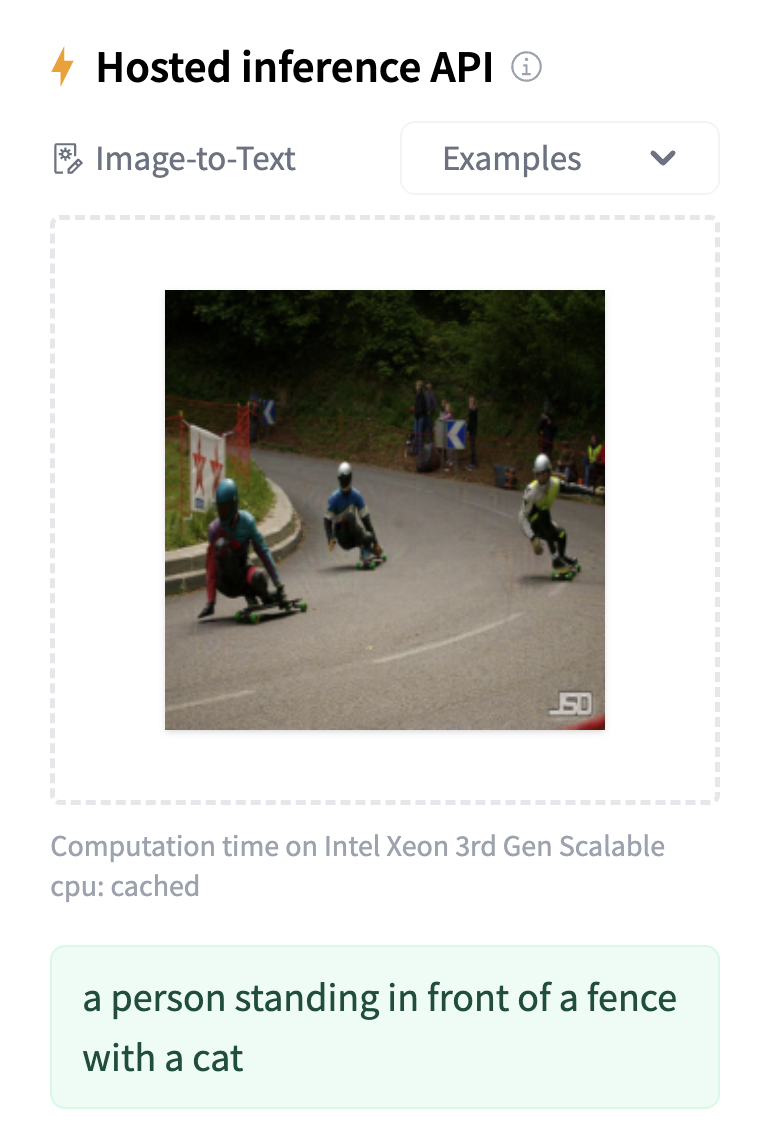}} & \quad
        \subfloat[(Orig)]{\includegraphics[width = 1.47in]{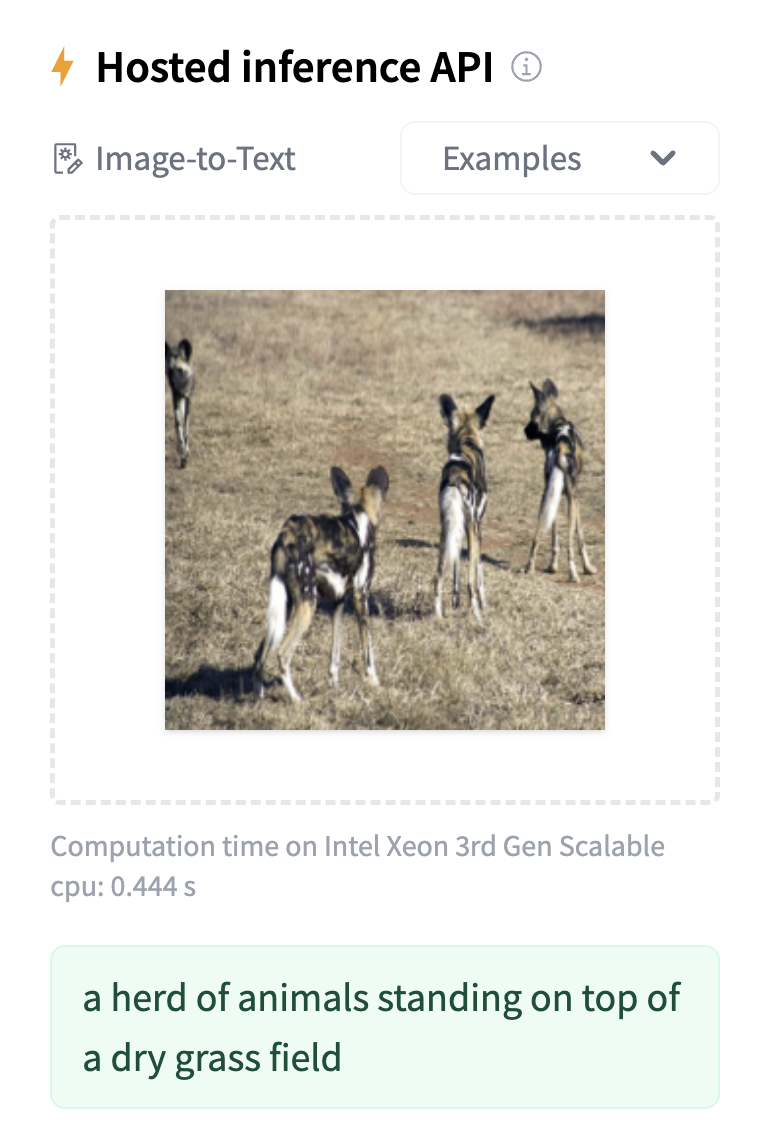}} &
        \subfloat[(Adv)]{\includegraphics[width = 1.47in]{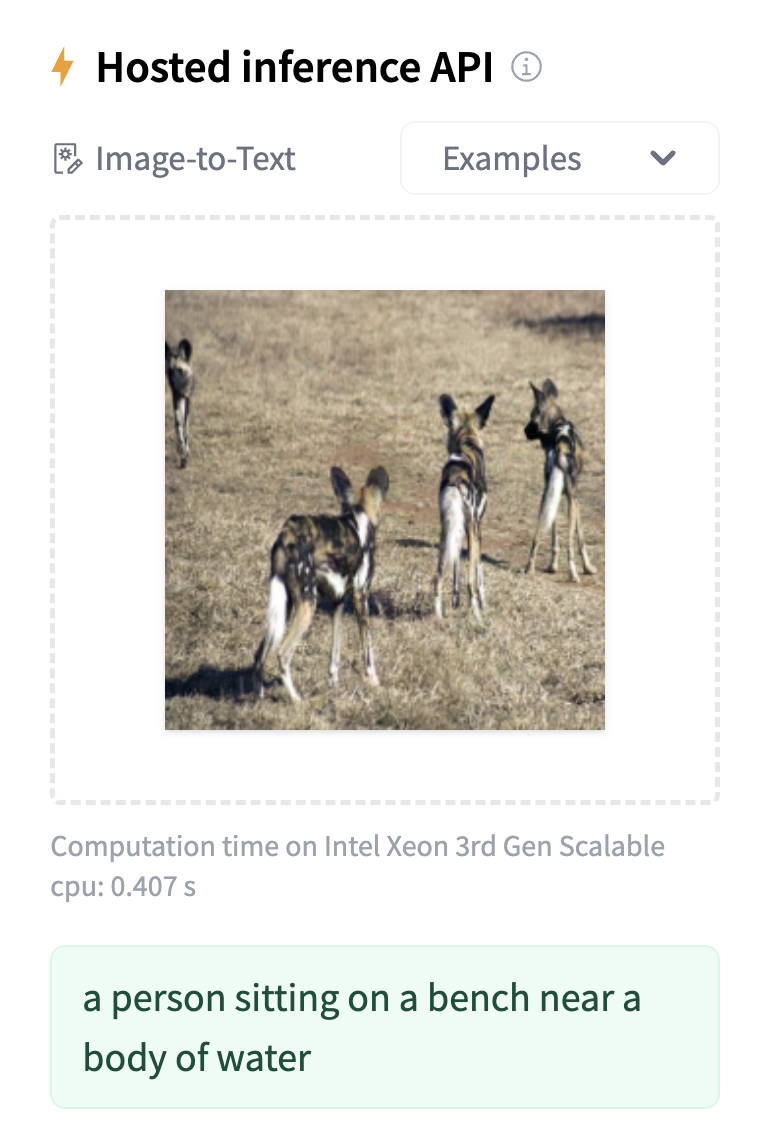}} \\
        
        \subfloat[(Orig)]{\includegraphics[width = 1.47in]{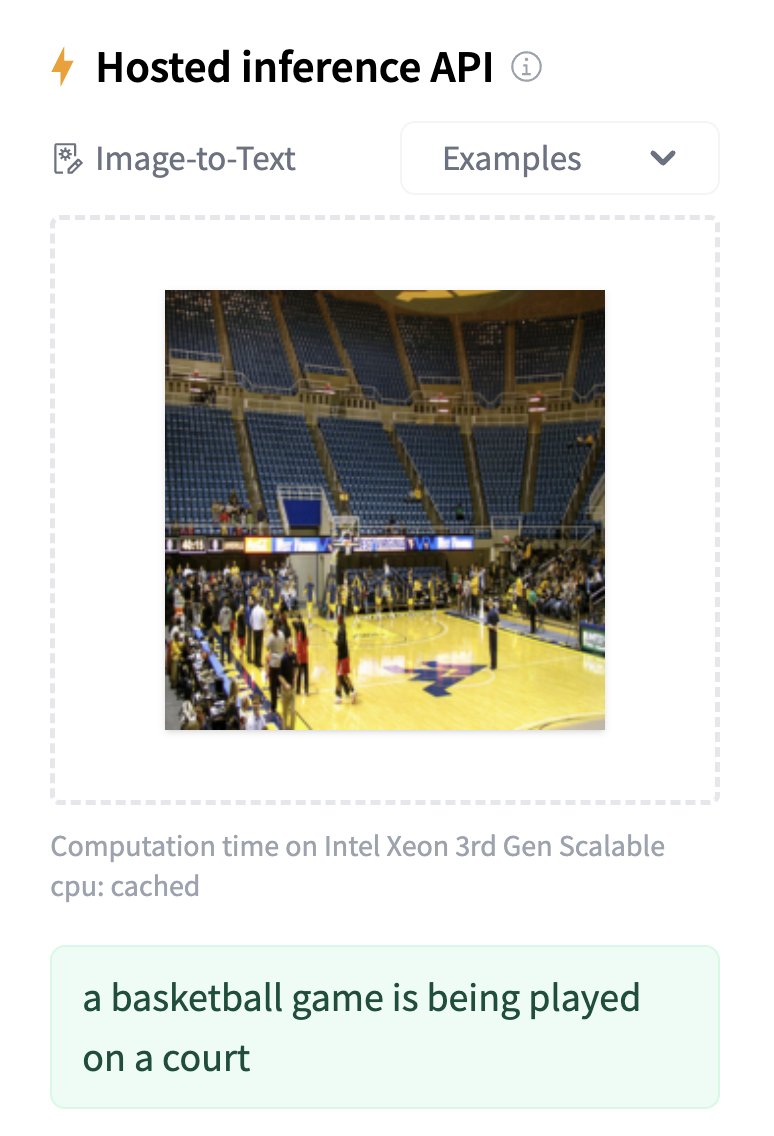}} &
        \subfloat[(Adv)]{\includegraphics[width = 1.47in]{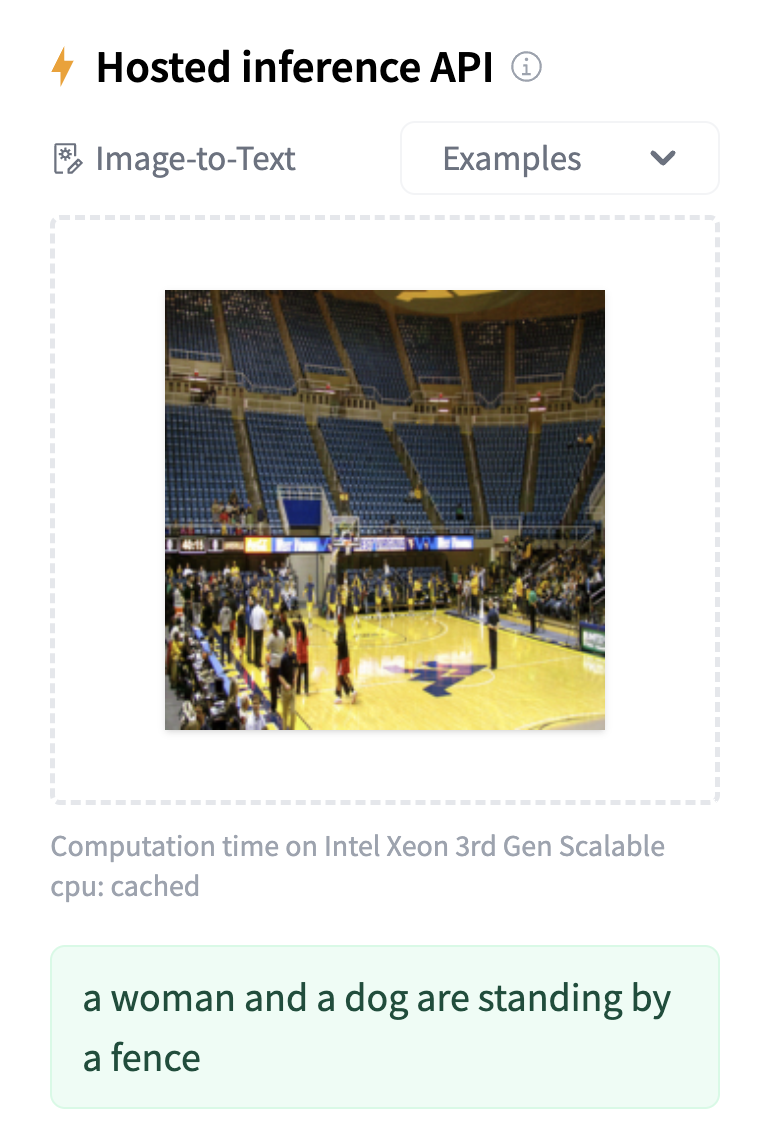}} & \quad
        \subfloat[(Orig)]{\includegraphics[width = 1.47in]{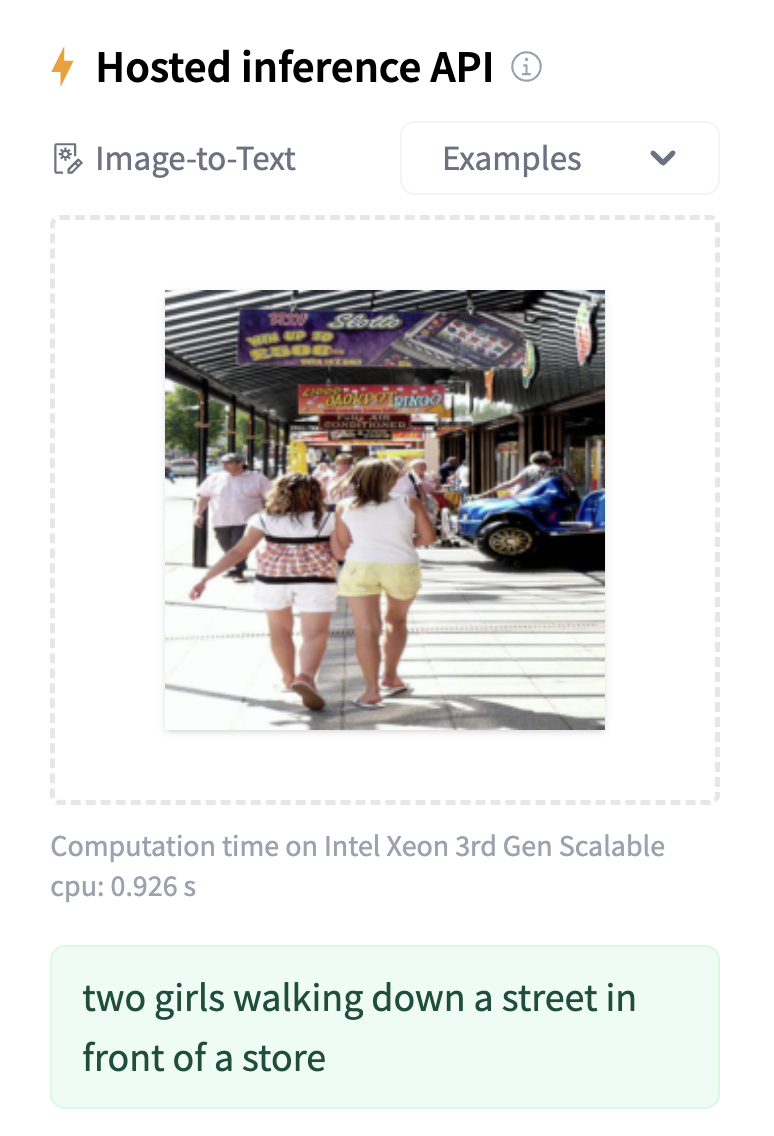}} &
        \subfloat[(Adv)]{\includegraphics[width = 1.47in]{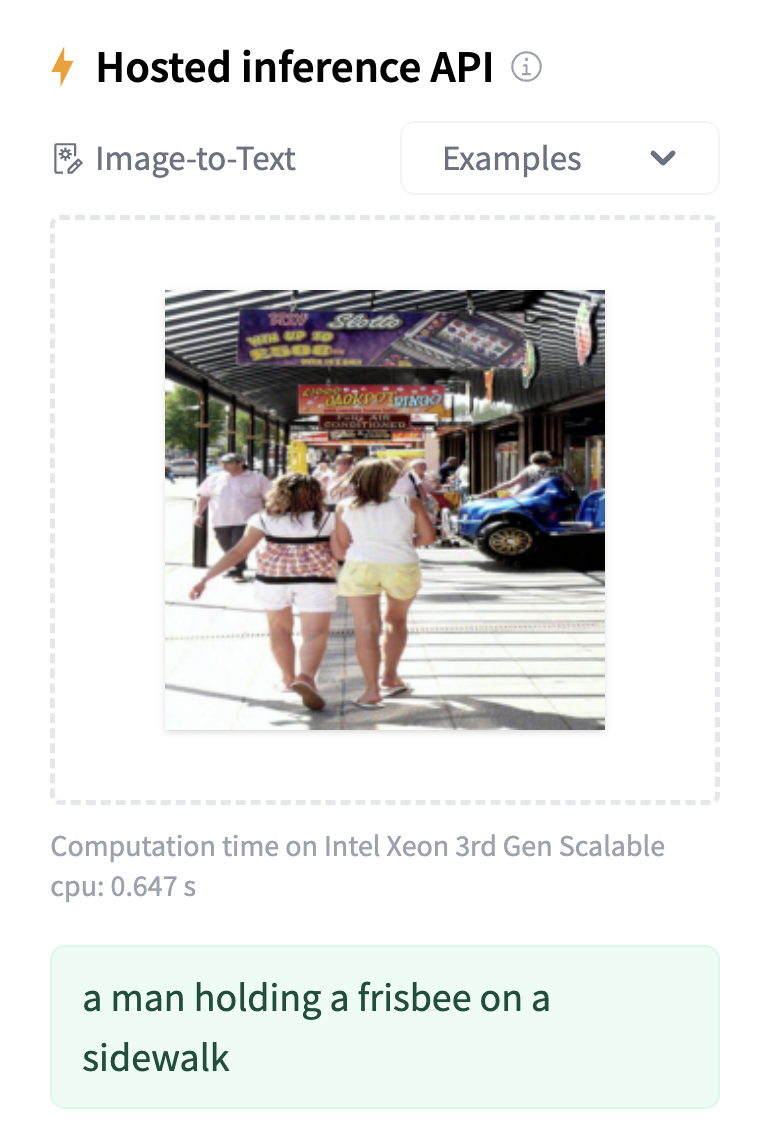}} \\
    \end{tabular}
    \caption{Original and adversarial images prediction using the hosted inference API by Hugging Face. Pairs of original images and their captions (Orig) beside adversarial images and their captions (Adv).}
    \label{images:huggingface}
\end{figure}

\textbf{Hugging Face results.} We generated our adversarial images and fed them through the Hugging Face API. The images succeeded completely in fooling the deployed model, even though we have no information on how it was deployed, nor whether it has been quantized or modified in the deployment procedure. \textbf{The same locally generated captions  were generated using the Hugging Face API}, thus making the local results shown in Table~\ref{tab:res} apply to the Hugging Face API. Hugging Face's API was unaware of the attack, because we were able to use the image encoder, and ``transfer'' the results optimized by us locally to the deployed model. Samples from the experiments are shown in Figure~\ref{images:huggingface}.

\section{Conclusions}
\label{sec:conclusions}
In this research, we made use of the vulnerability of the pretrained image encoder to input perturbations in order to develop a gray-box adversarial attack on the ViT-GPT2 model for image-to-text generation. Notably, our algorithm performs a gray-box attack without relying on specific language module information. 

The experiments demonstrated that our fairly simple method achieves a high success rate in generating adversarial examples. Further, we have shown that the proposed attack successfully fools the Hugging Face hosted inference API. Our experimental findings demonstrate that a carefully crafted human-imperceptible perturbation could effectively target image-to-text models. 

The widespread adoption of pretrained models brings new concerns regarding adversarial attacks, which exploit vulnerabilities and deceive the models into producing incorrect outputs. The robustness of these models need to be further analyzed.

\bibliographystyle{unsrtnat}
\bibliography{refs}

\end{document}